%% file: main.tex
\documentclass[10pt,conference,compsocconf]{IEEEtran}

\usepackage{hyperref}
\usepackage{graphicx}	
\usepackage{subcaption}
\usepackage{caption} 
\usepackage{xcolor}
\usepackage{forest}

\begin{document}
\title{Overcoming catastrophic forgetting in
 neural networks}

\author{
  Brandon Shuen Yi Loke, Filippo Quadri, Gabriel Vivanco, Maximilian Casagrande, Saúl Fenollosa \\
  \textit{École Polytechnique Fédérale de Lausanne}
}

\maketitle

\input{parts/abstract}
\input{parts/Introduction}
\input{parts/modelsandmethods}
\input{parts/setup}
\input{parts/results}

\input{parts/discussion}

\input{parts/summary}

\bibliographystyle{IEEEtran}
\bibliography{bib}

\input{parts/code_appendix}

\end{document}

%% file: parts/abstract.tex
\begin{abstract}

Catastrophic forgetting is the primary challenge that hinders continual learning, which refers to a neural network ability to sequentially learn multiple tasks while retaining previously acquired knowledge.
Elastic Weight Consolidation, a regularization-based approach inspired by synaptic consolidation in biological neural systems,  has been used to overcome this problem. In this study prior research is replicated and extended by evaluating EWC in supervised learning settings using the PermutedMNIST and RotatedMNIST benchmarks. Through systematic comparisons with L2 regularization and stochastic gradient descent (SGD) without regularization, we analyze how different approaches balance knowledge retention and adaptability. Our results confirm what was shown in previous research, showing that EWC significantly reduces forgetting compared to naive training while slightly compromising learning efficiency on new tasks. Moreover, we investigate the impact of dropout regularization and varying hyperparameters, offering insights into the generalization of EWC across diverse learning scenarios. These results underscore EWC’s potential as a viable solution for lifelong learning in neural networks.
\end{abstract}

%% file: parts/Introduction.tex
\section{Introduction}

An open challenge in Machine Learning is continuous learning, where a model gradually learns successive tasks without forgetting the previous ones. The main obstacle is therefore \textit{catastrophic forgetting}, which is the tendency of neural networks to lose performance on previously learned tasks when learning new ones \cite{French_1999}. This occurs when weights learned on earlier tasks are rewritten when trained on a different subsequent task. Overcoming this problem is essential for the development of scalable and robust systems capable of adapting to dynamic environments.

\begin{figure}[h!]
    \centering
    \includegraphics[width=0.7\linewidth]{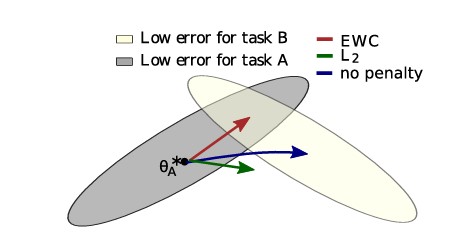}
    \caption{Schematic representation of EWC principle: learning a specific task establishes a range for each parameter of interest, within which it can vary without causing forgetting. Acquiring a new task requires adjusting the parameters to align with the overlapping ranges of previously learned tasks. \cite{Kirkpatrick_2017}}
    \label{fig:ewc_principle}
\end{figure}

The seminal work by James Kirkpatrick et al. \cite{Kirkpatrick_2017} introduced Elastic Weight Consolidation (EWC), a method inspired by neurobiological synaptic consolidation, to address catastrophic forgetting. EWC selectively reduces the plasticity of crucial weights for prior tasks, ensuring that learning the new ones minimally disrupts previous knowledge (Figure \ref{fig:ewc_principle}). The original study demonstrate the effectiveness of EWC using the PermutedMNIST dataset and Atari games, with a primary focus on reinforcement learning scenarios. 

In this project, EWC study is extended to supervised learning by reproducing and comparing the results from Figure 2A and 2B of the original paper \cite{Kirkpatrick_2017} on PermutedMNIST and RotatedMNIST. The reported analysis contrasts EWC performance with naive L2 regularization and no regularization, providing therefore insights into the relative strengths and weaknesses of each method.

This work aims to evaluate EWC's effectiveness in mitigating catastrophic forgetting and its potential as a general solution for sequential task learning in supervised settings.

The results obtained are discussed in Sec. \ref{sec:discussion}. Additional experiments, which were not part of the original paper, are presented in the appendix, along with a description of the code structure used.

%% file: parts/modelsandmethods.tex
\section{Models and Methods}

\subsection{Neural Network Architecture}
A fully connected feed-forward network (FCN) is adopted, featuring two hidden layers of 400 neurons each and ReLU activation. Let \(\mathbf{x}\in \mathrm{R}^{784}\) represent a flattened \(28\times28\) MNIST image, and let 
\(\theta = \{\mathbf{W}^{(1)}, \mathbf{b}^{(1)}, \mathbf{W}^{(2)}, \mathbf{b}^{(2)}, \mathbf{W}^{(3)}, \mathbf{b}^{(3)}\}\)
be the parameters, with the layer denoted in the superscript. The forward pass is given by:

{\footnotesize
\begin{equation}
\mathbf{h}_1 
= \mathrm{ReLU}\bigl(\mathbf{W}^{(1)} \mathbf{x} + \mathbf{b}^{(1)}\bigr),\quad
\mathbf{h}_2 
= \mathrm{ReLU}\bigl(\mathbf{W}^{(2)} \mathbf{h}_1 + \mathbf{b}^{(2)}\bigr)
\end{equation}
}
\begin{equation}
\mathbf{\hat{y}}
= \mathbf{W}^{(3)} \mathbf{h}_2 + \mathbf{b}^{(3)}
\end{equation}
where \(\mathbf{\hat{y}}\) is the logit output over the 10 classes. Training proceeds by minimizing the cross-entropy loss between the predicted distribution and the true labels. It is unclear in the original paper if batch normalization is used. Nevertheless, we include it in our model to enhance the network's performance. This is a technique that normalizes a layer's inputs by recentering and rescaling.

Where dropout is used in reproducing the figures, a probability of 0.2 was applied to the input layer and a probability of 0.5 applied to the hidden layers --- inline with the parameters used in the paper.

Early stopping was also implemented just as the authors did. If the validation error during cross-validation increased for more than five continuous iterations, the training is halted and the next dataset is used. The weights used are those computed before the rise in five continuous increases in validation error.

In general, parameters specified in the paper are also used in the study. When it is unclear what the original authors' chosen parameter is, cross-validation was performed to select the optimal parameter that minimizes the validation error.  

The original paper proposed a number of epochs per task ranging from 20 to 100. To evaluate the impact of training duration, experiments were conducted using both the same range of epochs and different values.

\subsection{Continual Learning Tasks}
\begin{figure}[b]
\includegraphics[width=0.3\textwidth]{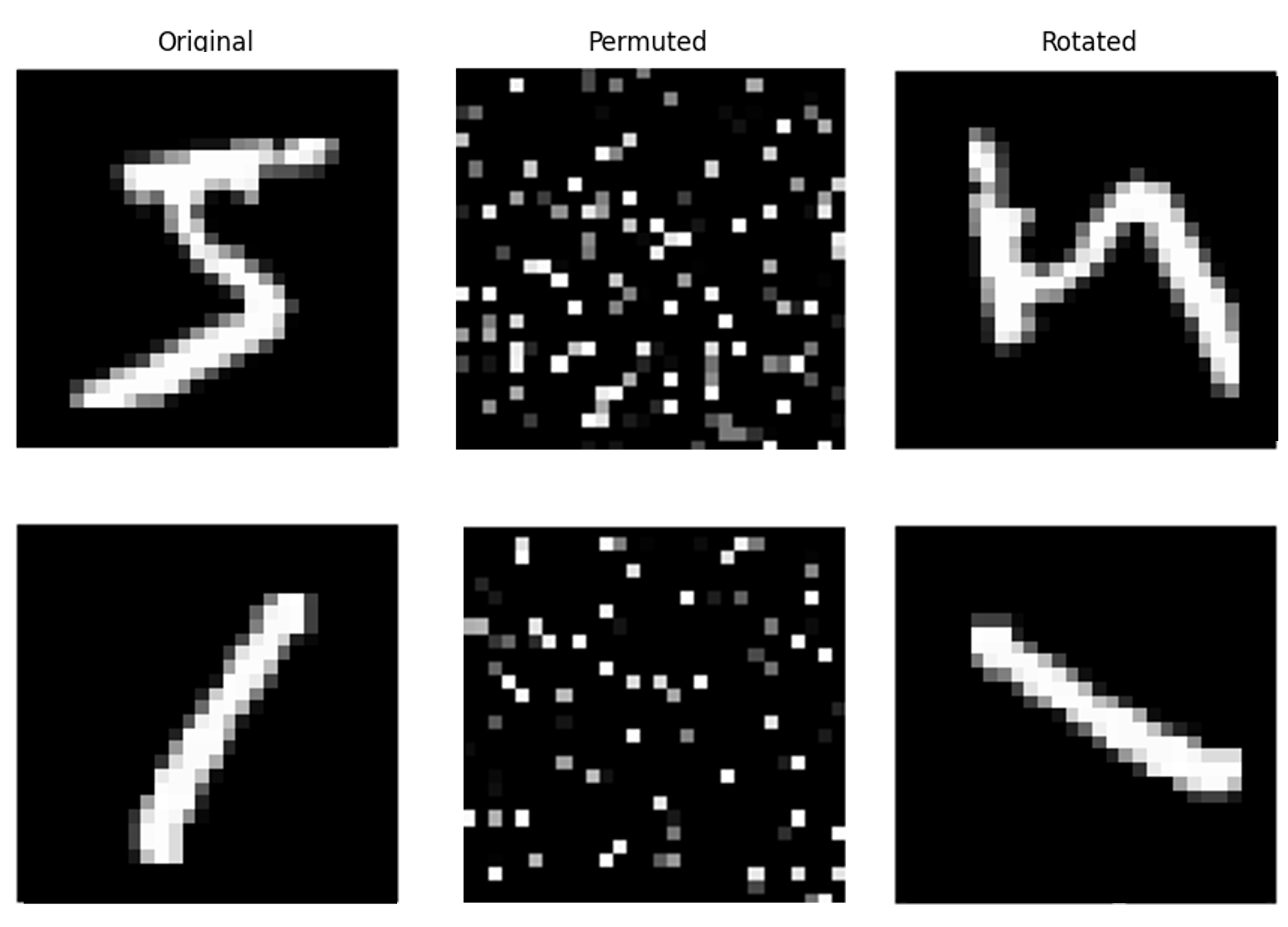}
\centering
\caption{\\
Visualization of dataset transformations: The first column shows the original MNIST digits. The second column presents the same digits after applying a fixed random permutation to the pixels (PermutedMNIST). The third column displays the digits rotated by a 50° (RotatedMNIST).}
\label{mnist}
\end{figure}

Two benchmarks for catastrophic forgetting are considered:
\begin{itemize}
    \item Permuted MNIST: Each task permutes the pixels of the MNIST images in a unique, fixed manner, as seen in Figure \ref{mnist}. Formally, each task \(t\) is defined by a permutation \(\pi_t\), and each sample \(\mathbf{x}\) is transformed into \(\mathbf{x}'\) according to \(\pi_t\). This benchmark was popularized in the context of continual learning by \cite{Goodfellow_2015}.
    \item Rotated MNIST: Each task rotates MNIST digits by a fixed angle, as shown in Figure \ref{mnist}. Denoting by \(R_{\alpha_t}\) the rotation by angle \(\alpha_t\), each image \(\mathbf{x}\) is replaced by \(R_{\alpha_t}(\mathbf{x})\).
\end{itemize}
Tasks are presented in a fixed sequence, and the model is trained on each task in turn.

\subsection{Regularization Approaches}

\paragraph{Naive Training (SGD)}
A baseline model is trained sequentially on each task with standard stochastic gradient descent (SGD) and momentum. No mechanism is introduced to preserve older tasks knowledge, which typically results in significant catastrophic forgetting \cite{Goodfellow_2015},

\paragraph{L2 Regularization}
A simple approach to mitigate forgetting involves applying an L2 penalty on parameter changes. If \(\mathcal{L}_{t}(\theta)\) is the cross-entropy loss for task \(t\), and \(\theta^{*}_A\) are the parameters optimized for task \(A\), then for a subsequent task \(B\):
\begin{equation}
\label{eq:l2}
\mathcal{L}(\theta)= \mathcal{L}_{B}(\theta) 
+ \frac{\lambda_{\mathrm{L2}}}{2}\|\theta-\theta^{*}_A\|^{2},
\end{equation}
where \(\lambda_{\mathrm{L2}} > 0\) balances the constraint.

\paragraph{Elastic Weight Consolidation (EWC)}
EWC improves on L2 by weighting each parameter's penalty according to the Fisher information matrix \(F\). Let \(F_{A}\) be computed after training task \(A\), approximated diagonally. The EWC loss on task \(B\) is:
\begin{equation}
\label{eq:ewc}
\mathcal{L}(\theta)
= \mathcal{L}_{B}(\theta)
+ \frac{\lambda_{\mathrm{EWC}}}{2}\sum_{i}F_{A,i}\bigl(\theta_{i}-\theta^{*}_{A,i}\bigr)^{2}.
\end{equation}
Parameters with higher \(F_{A,i}\) are penalized more strongly, thus preserving task \(A\). The generalization of the formula for $t$ tasks is
\begin{equation}
\label{eq:ewc_gen}
\mathcal{L}(\theta)
= \mathcal{L}_{T+1}(\theta)
+ \frac{\lambda_{\mathrm{EWC}}}{2}\sum_{t}\sum_{i}F_{t,i}\bigl(\theta_{i}-\theta^{*}_{t,i}\bigr)^{2}.
\end{equation}

\subsection{Training and Cross Validation}
Training was performed using mini-batch SGD and momentum. Hyperparameters such as batch size, momentum, and \(\lambda\) values (for both L2 and EWC) are selected via cross validation, particularly focusing on batch sizes (32, 64, or 128), momentum values ($0 - 0.9$), and regularization coefficients \(\lambda\) ranging from $1\cdot 10^{-5}-1$ for the L2 case and from $1-1\cdot10^{4}$ for the EWC. The validation accuracy is monitored to select optimal configurations. After this, the model is trained sequentially on tasks, reporting test accuracy for each previously learned task to measure performance degradation (forgetting) and average accuracy (forward transfer).

%% file: parts/setup.tex
\section{Experimental Setup}

This section details the findings from three principal experiments: (i) hyperparameter optimization through cross-validation, (ii) sequential learning on permuted MNIST, and (iii) sequential learning on rotated MNIST. The results highlight the magnitude of catastrophic forgetting across different regularization schemes and quantify how effectively each scheme preserves knowledge from earlier tasks.

\subsection{Cross-Validation}
The cross-validation results for both standard SGD (without regularization) and L2-regularized SGD showed that a batch size of 32 and a momentum of 0.6 provided a good balance between stable convergence and efficient training, achieving approximately 95\% accuracy on validation test sets. For L2 regularization, the best performance was observed with \(\lambda_{L2} \approx 0.01\). Experiments with higher values of \(\lambda_{L2}\) occasionally led to numerical instabilities, such as loss divergence or `NaN' losses.  
In the case of EWC, the optimal batch size was found to be 16, with a momentum of 0.2 and \(\lambda_{EWC} = 1000\). Here again, excessively high values of \(\lambda_{EWC}\) sometimes resulted in loss divergence during training.  

For the final experiments replicating the first figure, a batch size of 64 was used, with a momentum of 0.6 for both SGD and L2, and 0 for EWC, along with \(\lambda_{L2} = 0.01\) and \(\lambda_{EWC} = 10000\) (or \(\lambda_{EWC} = 20000\) depending on the task). For the second figure, additional cross-validation was performed on the \textit{SGD - Dropout} case, tuning the learning rate and hidden layer size. The best-performing hyperparameters were found to be a learning rate of \(1 \times 10^{-3}\) and a hidden layer width of $800$.  
It is important to note that these parameters may not be the absolute optimal ones but offer a reasonable trade-off between runtime and accuracy.

\subsection{Permuted MNIST}
For the permuted MNIST benchmark, each task was created by applying a distinct random permutation to the pixels of the original MNIST images. The different models -- plain stochastic gradient descent (SGD), L2-penalized SGD, and elastic weight consolidation (EWC) -- were trained on each new permutation sequentially. At the end of each task’s training, the accuracy on previously learned permutations was re-evaluated to measure forgetting. Performance metrics include per-task accuracy curves during training and the overall average accuracy across all tasks to illustrate each method’s capacity to maintain previously acquired knowledge. For the replication of the first figure, all models were evaluated on three permutations, whereas for the second figure, only SGD and EWC (with and without dropout) were tested across ten tasks.

\subsection{Rotated MNIST}
In the rotated MNIST setting, each task was generated by rotating the MNIST digits by an increment of \(10^\circ\). A total of ten rotations, from \(0^\circ\) to \(90^\circ\), were used to produce ten sequential tasks. As in the permuted case, models were trained on each new rotation and then tested on all previously learned rotations, facilitating a direct assessment of forgetting. Data was collected on task-specific accuracy throughout training, as well as the average performance over the entire set of tasks. Comparing these metrics for each regularization approach elucidates the extent to which rotation-induced variability exacerbates or mitigates catastrophic forgetting. Here again, for the replication of the first figure, all models were evaluated on three permutations, whereas for the second figure, only SGD and EWC (with and without dropout) were tested across ten tasks.

\subsection{Expected Patterns}
The experiments are designed to observe the progressive decay in accuracy on older tasks as each subsequent task is introduced. Plain SGD typically exhibits more pronounced catastrophic forgetting, whereas using L2 regularization is expected to partially constrain parameter drift and thus retain moderate performance on old tasks. However, this comes at the cost of encountering serious problem in learning new ones. EWC, which prioritizes parameters deemed critical for prior tasks via the Fisher information, is hypothesized to mitigate forgetting more effectively. The collected plots present task-wise performance and overall averaged performance at each training phase, illustrating how well each approach balances learning of new data with retention of prior knowledge.

%% file: parts/results.tex
\section{Results}

\begin{figure*}
    \centering
    \begin{subfigure}{0.48\textwidth} 
        \centering
        \includegraphics[height=5.5cm]{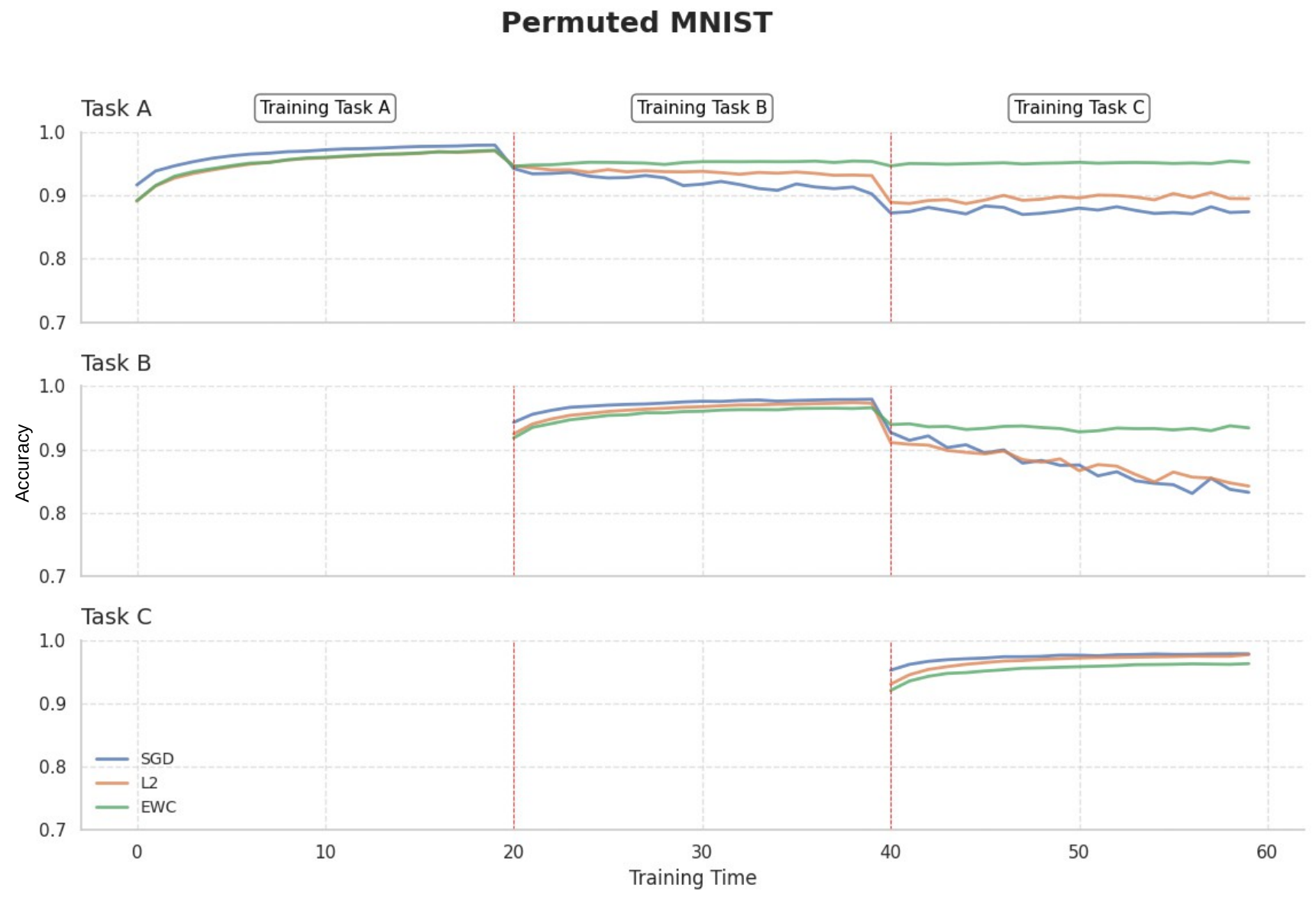}
        \label{fig:permuted_first_fig}
    \end{subfigure}
    \hfill
    \begin{subfigure}{0.5\textwidth} 
        \centering
        \vfill 
        \includegraphics[height=5cm]{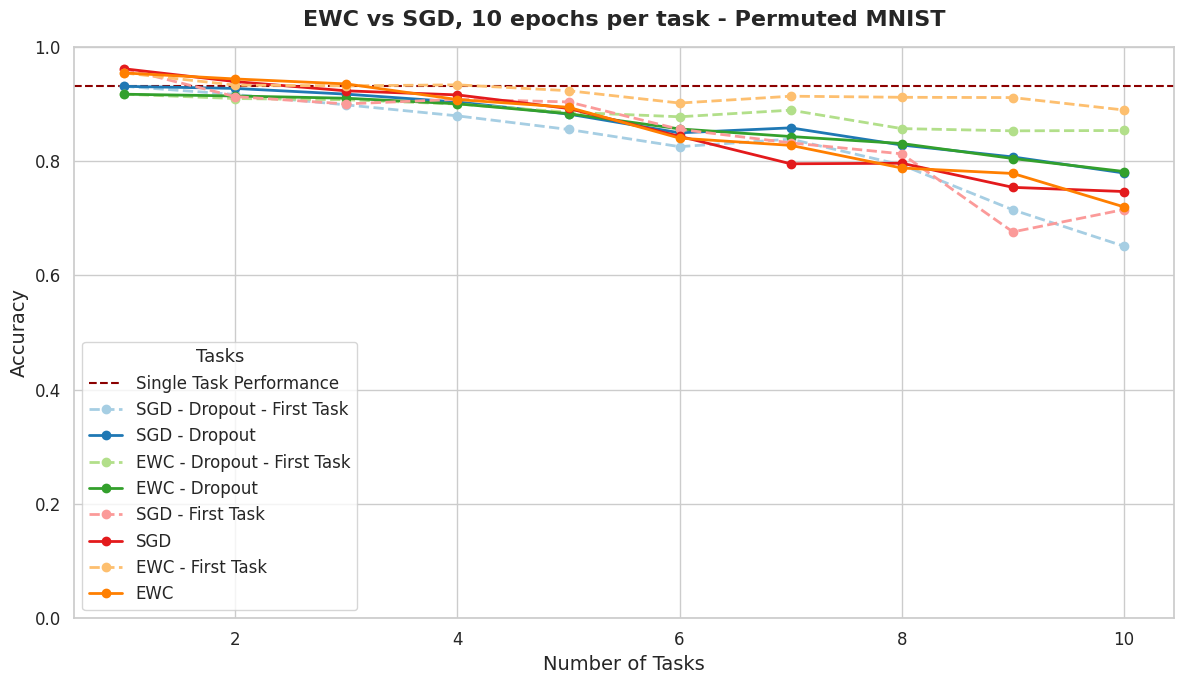}
        \label{fig:permited_second_fig}
    \end{subfigure}

    \captionsetup{justification=justified, singlelinecheck=false}

    \caption{\\Custom reproduction of the two figures from the original paper \cite{Kirkpatrick_2017}. The left figure compares the performance of \textit{SGD}, \textit{SGD + L2 regularization}, and \textit{SGD + EWC regularization} on three permutations. The right figure presents a similar comparison across 10 different tasks, evaluating \textit{SGD} and \textit{EWC models} (with and without dropout). As a reference, \textit{Single Task Performance} is computed, corresponding to the accuracy on the Task 1 test set after training with the \textit{SGD + Dropout model}.}
    \label{fig:perm_res}
\end{figure*}

As previously described, the experiments from the original paper \cite{Kirkpatrick_2017} were replicated on two datasets: \textit{PermutedMNIST} (as in the original study) and \textit{RotatedMNIST}. The results for \textit{PermutedMNIST} are presented in Figure \ref{fig:perm_res}, while those for \textit{RotatedMNIST} are provided in the \textit{Appendix} section, on Figure \ref{fig:rotated_mnist} and Figure \ref{fig:seq_rot_10}. Additional variations of the original experiments were conducted, including an evaluation of \textit{EWC with dropout regularization}, an evaluation for a \textit{Mixed Dataset}, a comparison of \textit{10-task performance} using different numbers of epochs, and the implementation of early stopping. All corresponding figures are available in the \textit{Appendix} section, Figure \ref{fig:benchmark_comparison} and Figure \ref{fig:seq_learning}.

%% file: parts/discussion.tex
\section{\label{sec:discussion}Discussion}

Let's begin by analyzing the three-task comparison (Appendix, Figure \ref{fig:benchmark_comparison}) for four case studies: \textit{Permuted MNIST}, \textit{Rotated MNIST} (0°, 40° and 90°), \textit{Mixed MNIST} (Rotated 0°, Permuted, Rotated 90°), and \textit{Permuted MNIST - Dropout}, where a \textit{Dropout regularization} is added to the networks. Dropout regularization is a technique used to prevent overfitting in neural networks. During training, random neurons are ``dropped out" (set to zero) at each iteration with a certain probability. This forces the network to rely on multiple pathways, making it more robust and less likely to overfit to the training data. 

It can generally be observed that EWC regularization better prevents catastrophic forgetting, particularly for Task A. However, the improved maintenance of performance on Task A sometimes results in slightly lower learning performance on subsequent tasks. Another interesting observation is that the permuted tasks exhibit less forgetting compared to the rotated ones, likely because the permutation only requires learning a pattern, while the rotation (with a big difference in angle from a task to another) involves more complex transformations that change the input significantly across tasks. An additional experiment with \textit{Rotated MNIST} at angles of 0°, 10°, and 20° showed nearly the same performance as in \textit{Permuted MNIST} (Appendix, Figure \ref{fig:rotated_mnist_small}). The \textit{Permuted MNIST - Dropout} and \textit{Mixed Task} cases are interesting. Dropout learns in a slightly worse way than the simple permutation due to fewer neurons being used. In this case, the EWC regularization reduces forgetting by leaving more ``elastic" neurons for later tasks. In contrast, SGD and L2 regularizations cause a significant drop in Task A performance when learning Task C. This happens because they tightly constrain the model to the task they are learning, making it harder to retain knowledge from earlier ones, leading to catastrophic forgetting. Also in the \textit{Mixed Task} case, the EWC performs better than the SDG and L2 cases, demonstrating its ability to generalize better across tasks and reduce catastrophic forgetting.

For the ten-task comparison (Appendix, Figure \ref{fig:seq_learning}), only the SGD method and EWC (with and without dropout regularization) were tested. The original paper recommends training for 100 epochs per task with early stopping. Early stopping is configured with a patience of 5 epochs, based on the validation set, which includes all the test sets of previously learned tasks. However, in our experiments, Figure \ref{fig:seq_perm_100} and \ref{fig:seq_rot_100}, it was observed that using this validation set, the majority of tasks were effectively trained for only two epochs. This early stopping wasn’t due to overfitting, but rather the forgetting of previously learned tasks. While the reduced number of effective tasks might be sufficient for the simple tasks we studied, it doesn’t fully reflect a real-world scenario. Thus, an additional test was run (Figure \ref{fig:seq_perm_10} and \ref{fig:seq_rot_10}) where each task was trained for a fixed 10 epochs without early stopping. The results show that catastrophic forgetting is less pronounced with EWC compared to SGD, particularly in terms of performance on the first task. As mentioned earlier, the \textit{Rotated MNIST} task appears harder to "remember", and a noticeable difference between the two methods in both the first task performance and overall performance (evaluated by combining the test sets of all previously learned tasks) can be observed. In contrast, for the \textit{Permuted MNIST} task, only the performance on the first task is significantly better for EWC. As shown in Figure \ref{fig:permuted_mnist}, all models perform well on this dataset, with less pronounced forgetting. Therefore, the relatively minor forgetting of previous tasks, coupled with strong learning of new tasks, may yield similar results to EWC.

%% file: parts/summary.tex
\section{Conclusion}
The numerical results demonstrate that Elastic Weight Consolidation is an effective regularization method for mitigating catastrophic forgetting. By selectively reducing the plasticity of certain parameters, EWC enables continual learning with minimal degradation of previously acquired knowledge. Our implementation successfully replicates the results from Kirkpatrick et al. \cite{Kirkpatrick_2017}, particularly on the \textit{PermutedMNIST} benchmark, where EWC outperforms SGD and L2 regularization.
Despite these promising results, our implementation does not fully match the accuracy reported in the original paper. This discrepancy could stem from differences in hyperparameter tuning, network architecture, or computational constraints that limited our ability to explore a broader parameter space. Furthermore, the near-perfect performance reported raises questions about potential implicit biases or undisclosed training tricks.
From a broader perspective, the ability to retain knowledge across sequential tasks has significant applications, particularly in lifelong learning systems, robotics, and adaptive AI. EWC can be instrumental in environments requiring such continual adaptation. However, its effectiveness is still constrained by the nature of the tasks.
Future work could explore hybrid approaches, combining EWC with alternative strategies, to further improve long-term retention while maintaining adaptability to new tasks. Additionally, applying these techniques to more complex real-world datasets beyond MNIST would provide further insights into their scalability and practical usability.

%% file: parts/code_appendix.tex
\onecolumn
\appendix

\section*{Benchmark Performance Comparisons}

\begin{figure}[h]
    \centering
    \begin{subfigure}{0.4\textwidth}
        \centering
        \includegraphics[width=\textwidth]{figures/permuted_minst.pdf}
        \caption{Accuracy curves for the \textit{PermutedMNIST} benchmark across three tasks, comparing standard SGD, L2 regularization, and EWC.}
        \label{fig:permuted_mnist}
    \end{subfigure}
    \begin{subfigure}{0.4\textwidth}
        \centering
        \includegraphics[width=\textwidth]{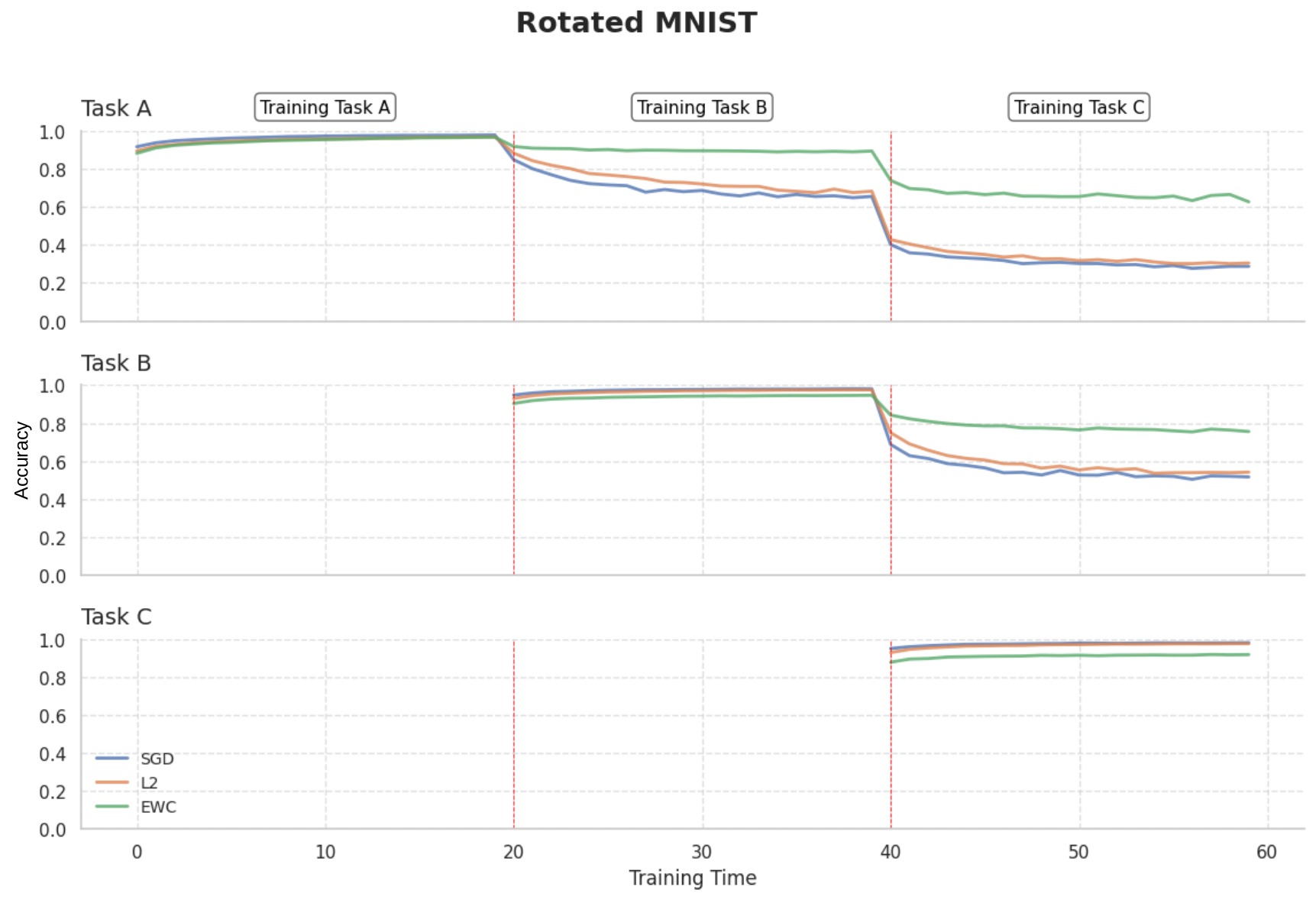}
        \caption{Accuracy curves for the \textit{RotatedMNIST} benchmark at 0°, 40°, and 90° rotation angles, showing performance degradation across tasks.}
        \label{fig:rotated_mnist}
    \end{subfigure}

    \vspace{0.5cm}

    \begin{subfigure}{0.4\textwidth}
        \centering
        \includegraphics[width=\textwidth]{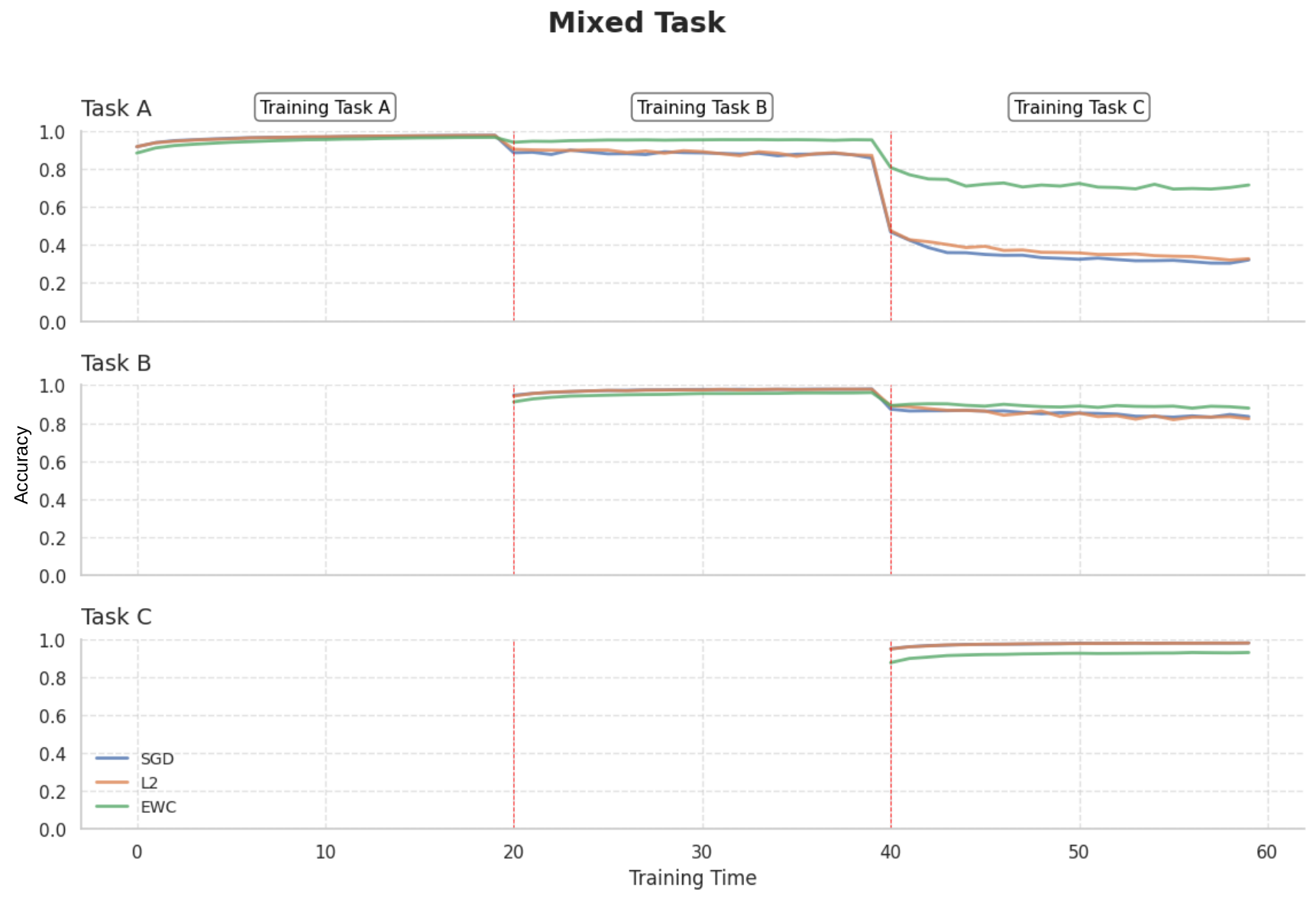}
        \caption{Results on a mixed benchmark (Rotated 0°, Permuted, Rotated 90°), demonstrating the relative effectiveness of EWC in heterogeneous task sequences.}
        \label{fig:mixed_mnist}
    \end{subfigure}
    \begin{subfigure}{0.4\textwidth}
        \centering
        \includegraphics[width=\textwidth]{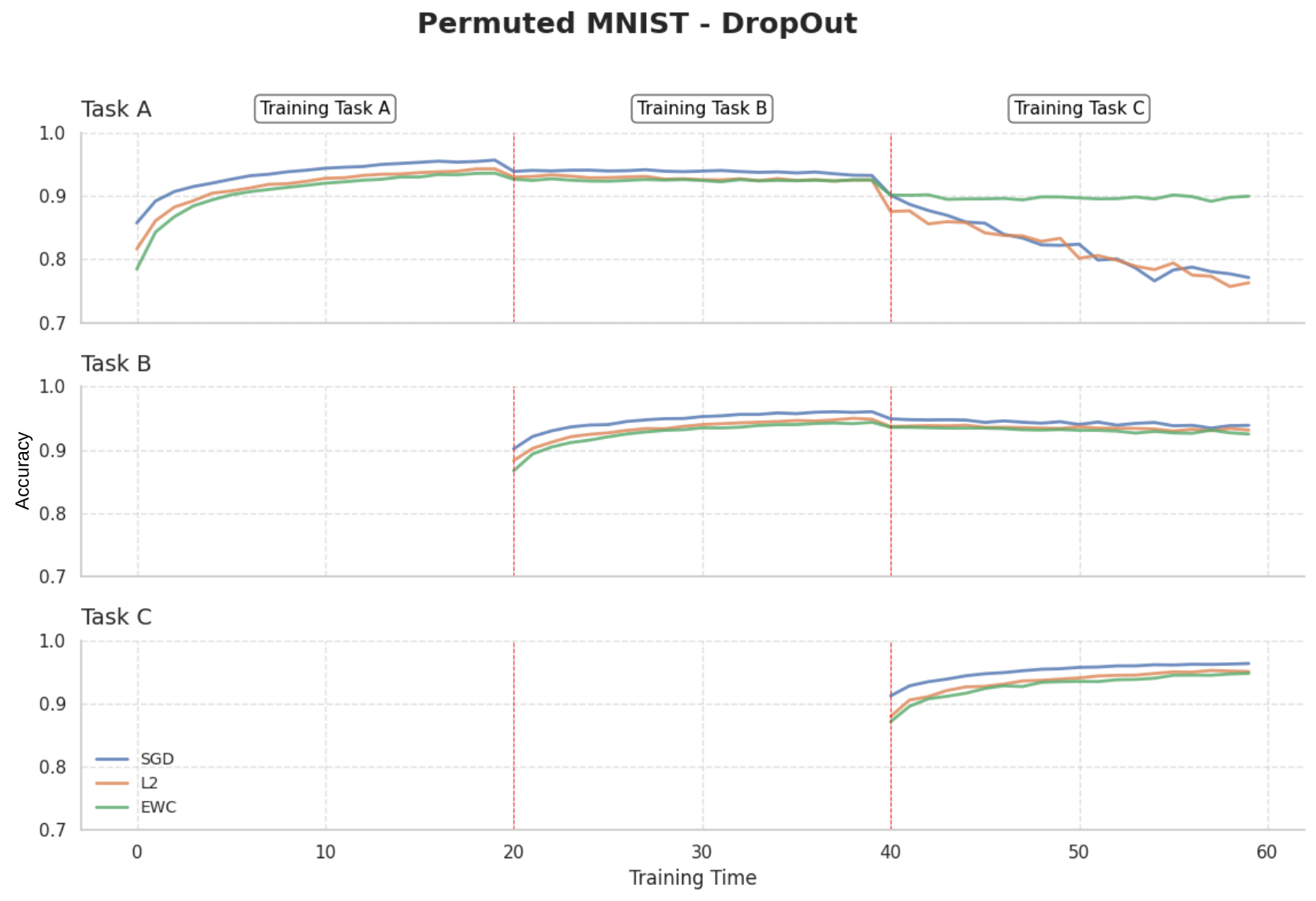}
        \caption{\textit{PermutedMNIST} with Dropout: Performance comparison of SGD, L2, and EWC when dropout regularization is applied, highlighting its effect on stability.}
        \label{fig:permuted_mnist_dropout}
    \end{subfigure}

    \vspace{0.5cm}

    \begin{subfigure}{0.4\textwidth}
        \centering
        \includegraphics[width=\textwidth]{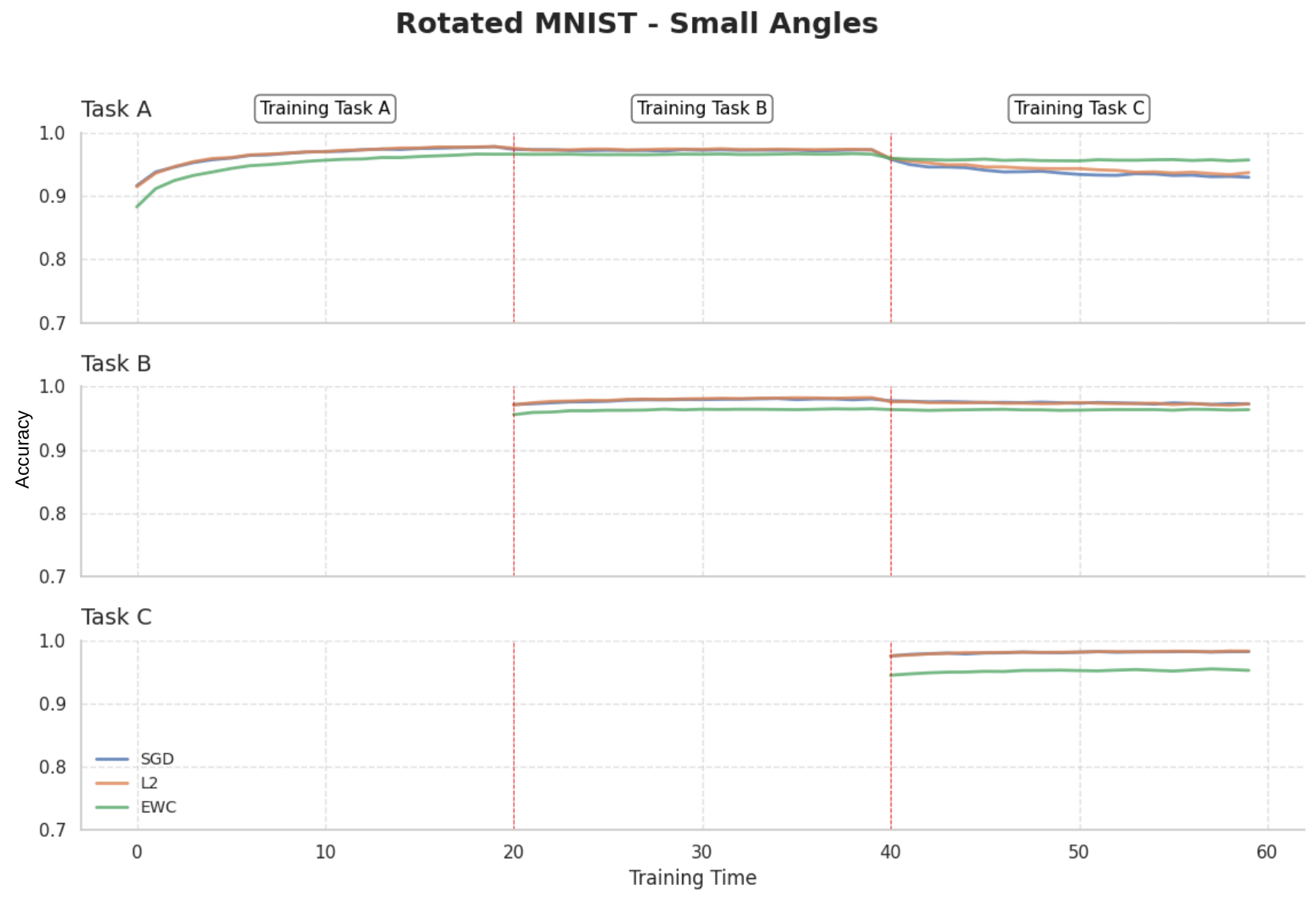}
        \caption{Accuracy curves for the \textit{RotatedMNIST} benchmark at smaller angles (0°, 10°, and 20°), showing notably less performance degradation across tasks.}
        \label{fig:rotated_mnist_small}
    \end{subfigure}
    
    \caption{Comparison of different regularization methods on three-task benchmarks.}
    \label{fig:benchmark_comparison}
\end{figure}

\clearpage

\section*{Sequential Task Learning Performance}
\bigskip
\bigskip

\begin{figure}[h]
    \centering
    \begin{subfigure}[t]{0.45\textwidth}  
        \centering
        \includegraphics[width=\textwidth]{figures/figure_2_perm_10_epochs.png}
        \caption{Sequential task learning results for SGD and EWC on \textit{PermutedMNIST} across ten tasks, with 10 epochs per task, showing the extent of catastrophic forgetting.}
        \label{fig:seq_perm_10}
    \end{subfigure}
    \hspace{1cm}  
    \begin{subfigure}[t]{0.45\textwidth}  
        \centering
        \includegraphics[width=\textwidth]{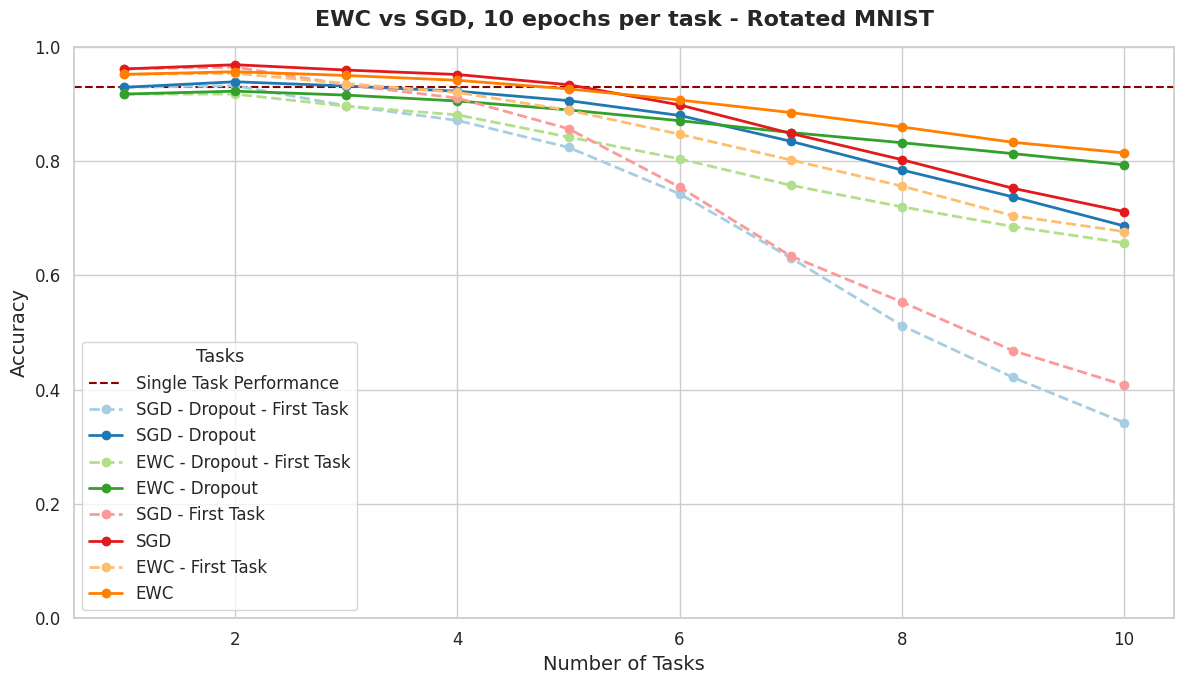}
        \caption{Sequential task learning results using \textit{RotatedMNIST}. 10 epochs per task were performed.}
        \label{fig:seq_rot_10}
    \end{subfigure}

    \vspace{1cm} 

    \begin{subfigure}[t]{0.45\textwidth}  
        \centering
        \includegraphics[width=\textwidth]{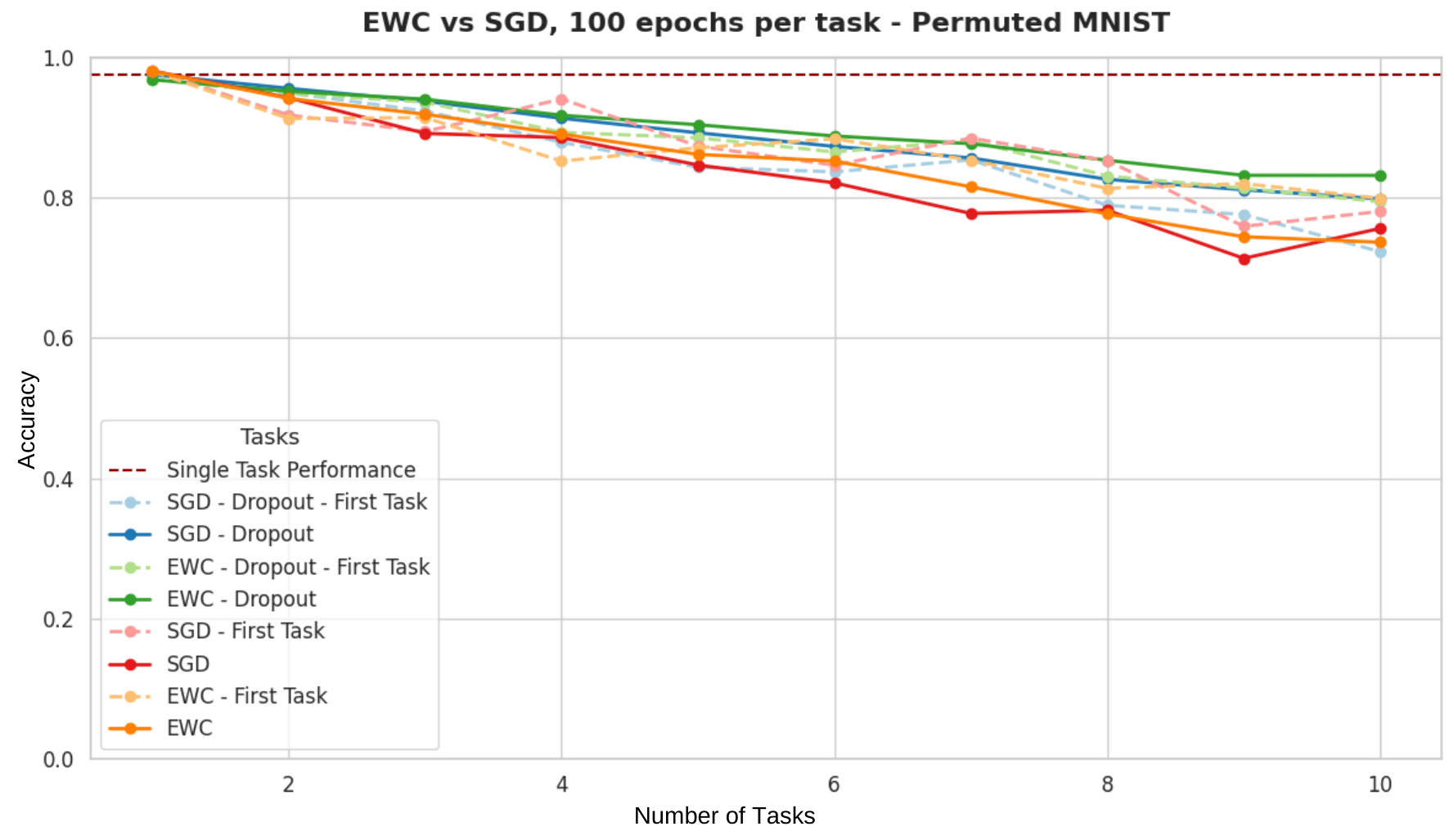}
        \caption{Results from Sequential task learning on \textit{PermutedMNIST} across ten tasks, with 100 epochs per task and early stopping.}
        \label{fig:seq_perm_100}
    \end{subfigure}
    \hspace{1cm}  
    \begin{subfigure}[t]{0.45\textwidth}  
        \centering
        \includegraphics[width=\textwidth]{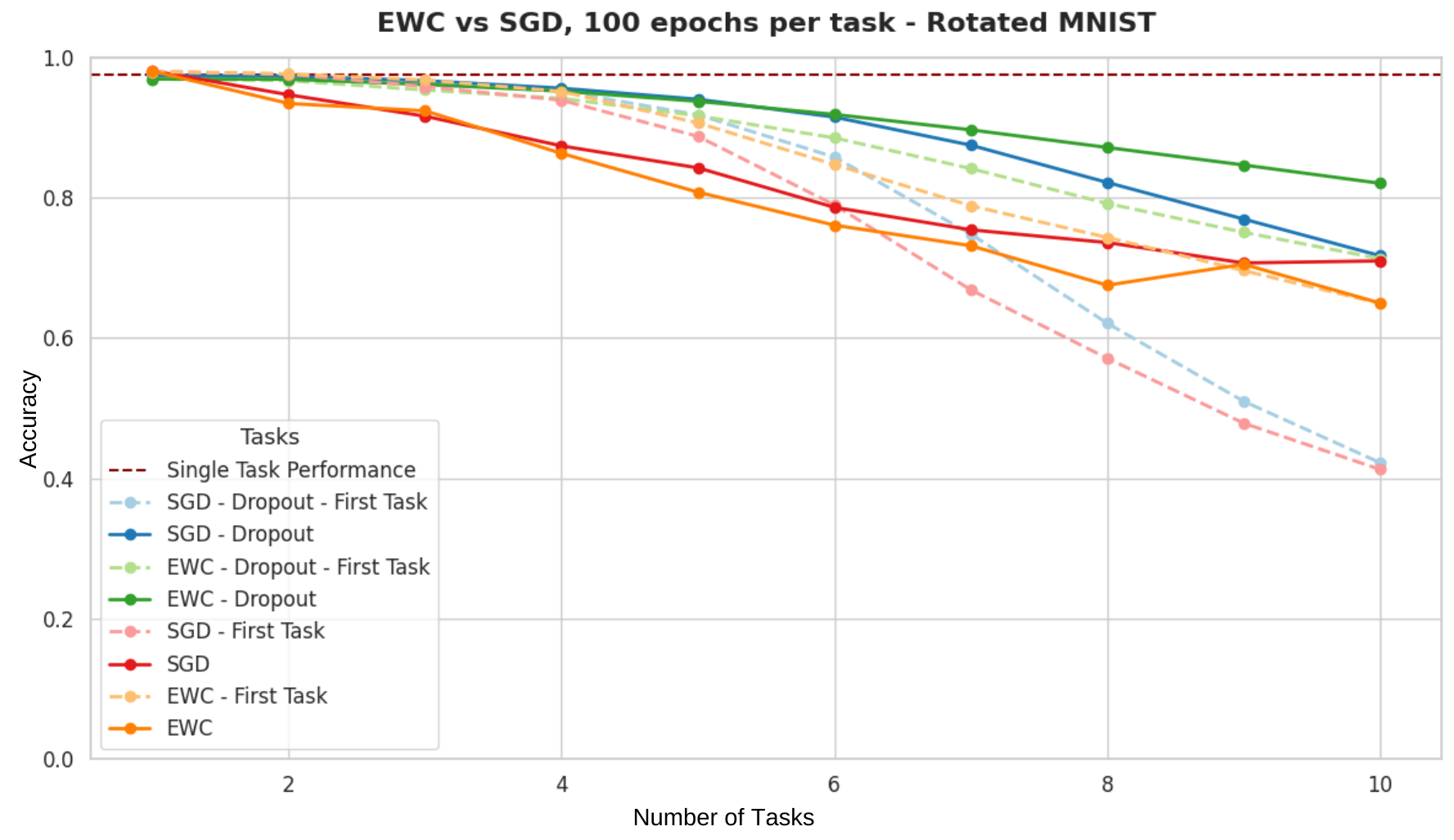}
        \caption{Sequential task learning results for SGD and EWC on \textit{RotatedMNIST}, where task complexity affects retention across methods. 100 epochs per task were performed and early stopping technique was applied.}
        \label{fig:seq_rot_100}
    \end{subfigure}

    \caption{Performance analysis on the ten-task benchmark.}
    \label{fig:seq_learning}
\end{figure}

\clearpage

\section*{Code Structure Overview}
\bigskip
\bigskip

The project includes a set of utility scripts and notebooks to facilitate dataset generation, model training, and evaluation. Below is an overview of the key components:

\subsection{Utility Scripts}
Located in the \texttt{utils} folder:
\begin{itemize}
    \item \texttt{data\_utils.py}: Utilities for dataset generation.
    \item \texttt{ewc.py}: Implementation of EWC regularization.
    \item \texttt{l2.py}: Definition of L2 regularization.
    \item \texttt{train\_utils.py}: Functions for training the models.
    \item \texttt{viz\_utils.py}: Functions for visualizing results.
\end{itemize}

\subsection{Notebooks}
\begin{itemize}
    \item \texttt{overcoming\_catastrophic\_forgetting\_cross\_val.ipynb}: Used for cross-validation to optimize model parameters.
    \item \texttt{overcoming\_catastrophic\_forgetting\_in\_NN.ipynb}: Used for building and training the fully connected neural network (from now referred as \texttt{NN\_notebook}).
\end{itemize}

\vspace{1cm}
\section*{Figure Reproduction}  
\bigskip  
\bigskip  

This section provides a brief explanation of how to reproduce the subfigures in Figure \ref{fig:benchmark_comparison} and Figure \ref{fig:seq_learning}.  

\subsection{Figure \ref{fig:benchmark_comparison}}  
The subfigures in Figure \ref{fig:benchmark_comparison} were generated using the first part of the \texttt{NN\_notebook}.  

\begin{itemize}  
    \item \textbf{Figure \ref{fig:permuted_mnist}:} All networks are of type \texttt{FCN}, and the data loaders were generated using \textit{Permuted MNIST}.  
    \item \textbf{Figure \ref{fig:rotated_mnist}:} All networks are of type \texttt{FCN}, and the data loaders were generated using \textit{Rotated MNIST} with rotation angles of 0°, 40°, and 90°.  
    \item \textbf{Figure \ref{fig:mixed_mnist}:} All networks are of type \texttt{FCN}, and the data loaders for the three tasks were: Rotated 0°, Permuted, and Rotated 90°.  
    \item \textbf{Figure \ref{fig:permuted_mnist_dropout}:} All networks are of type \texttt{FCN\_Dropout}, and the data loaders were generated using \textit{Permuted MNIST}.  
    \item \textbf{Figure \ref{fig:rotated_mnist_small}:} All networks are of type \texttt{FCN}, and the data loaders were generated using \textit{Rotated MNIST} with rotation angles of 0°, 10°, and 20°.  
\end{itemize}  

\subsection{Figure \ref{fig:seq_learning}}  
The subfigures in Figure \ref{fig:seq_learning} were generated using the second part of the \texttt{NN\_notebook}.  

\begin{itemize}  
    \item \textbf{Figures \ref{fig:seq_perm_10} and \ref{fig:seq_rot_10}:} The second part of the notebook was executed with the \texttt{train\_with\_avg\_perf} function configured to a patience of 15 and 10 epochs (without early stopping).  
    \item \textbf{Figures \ref{fig:seq_perm_100} and \ref{fig:seq_rot_100}:} The second part of the notebook was executed with the \texttt{train\_with\_avg\_perf} function configured to a patience of 5 and 100 epochs.  
\end{itemize}  

\vspace{1cm}
\section*{Task Distribution}  
\bigskip
\bigskip

The majority of the work was carried out collaboratively by the entire group. However, specific tasks were primarily handled by the following members:  
\begin{itemize}  
    \item \texttt{L2} and \texttt{SGD}: Saúl Fenollosa, Maximilian Casagrande  
    \item \texttt{EWC}: Filippo Quadri, Gabriel Vivanco  
    \item \texttt{Cross Validation}: Brandon Shuen Yi Loke  
    \item \texttt{Report}: All group members  
\end{itemize}